\pdfoutput=1

\documentclass[11pt]{article}
\usepackage{multirow}
\usepackage{EMNLP2023}

\usepackage{times}
\usepackage{latexsym}

\usepackage[T1]{fontenc}

\usepackage[utf8]{inputenc}

\usepackage{microtype}

\usepackage[textsize=footnotesize]{todonotes}

\usepackage{amsmath}
\usepackage{amssymb}
\usepackage{inconsolata}
\usepackage{graphicx}
\usepackage{subfigure}
\usepackage{caption}
\usepackage{subcaption}
\usepackage{booktabs}
\usepackage[utf8]{inputenc}
\title{Assessing the Reliability of Large Language Model Knowledge}
\begin{document}
\author{%
Weixuan Wang\textsuperscript{1}, Barry Haddow\textsuperscript{1}, Alexandra Birch\textsuperscript{1}, 
Wei Peng\textsuperscript{2}\\[1ex]
$^1$ School of Informatics, University of Edinburgh\\
\small
w.wang-126@sms.ed.ac.uk, bhaddow@ed.ac.uk, a.birch@ed.ac.uk \\
$^2$ Huawei Technologies Co., Ltd.\\
\small
peng.wei1@huawei.com
}

\maketitle
\begin{abstract}

Large language models (LLMs) have been treated as knowledge bases due to their strong performance in knowledge probing tasks. LLMs are typically evaluated using accuracy, yet this metric does not capture the vulnerability of LLMs to hallucination-inducing factors like prompt and context variability. How do we evaluate the capabilities of LLMs to consistently produce factually correct answers? In this paper, we propose \textbf{MOdel kNowledge relIabiliTy scORe (MONITOR)}, a novel metric designed to directly measure LLMs' factual reliability. MONITOR computes the distance between the probability distributions of a valid output and its counterparts produced by the same LLM probing the same fact using different styles of prompts and contexts.
Experiments on a comprehensive range of 12 LLMs demonstrate the effectiveness of MONITOR in evaluating the factual reliability of LLMs while maintaining a low computational overhead. In addition, we release the \textbf{FKTC} (Factual Knowledge Test Corpus) test set, containing 210,158 prompts in total to foster research along this line\footnote{https://github.com/Vicky-Wil/MONITOR}. 

\end{abstract}

\section{Introduction}

Recently large pre-trained language models (LLMs),
especially those with billions of parameters, have been used as de facto storage for factual knowledge. Applying LLMs to real-world scenarios inevitably leads to language generation deviating from known facts (aka ``factual hallucination'' \citep{survey}) due to multiple causes. For example, \citet{guess} argued that the performance of an LLM is over-estimated due to biased prompts over-fitting datasets (also referred to as the framing effect in \citet{cognitive}) and in-context information leakage. 

Given the variability of LLMs' performance under different prompts and contexts, it seems that purely evaluating them on accuracy is not enough and that we also need to gauge how robust they are to variations in prompting.
In Figure 1 we show examples of factual probes where either the framing of the prompt, or the context to the prompt, is varied, leading to the issue of ``accuracy instability''.  

\begin{figure}[ht]
\centering                                                          
\subfigure[Prompt framing effect]{\label{semantic-bias}\includegraphics[scale=0.5]{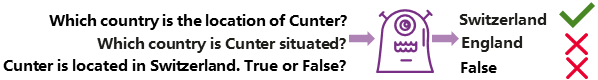}}        
\subfigure[Effect of in-context interference]{\label{leakage-bias}\includegraphics[scale=0.5]{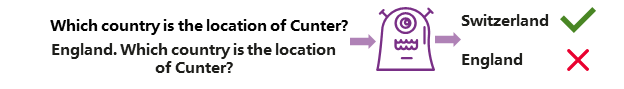}} 
\caption{``Accuracy instability'' during language generation under various prompts.}
\label{bias} 
\end{figure}


\paragraph{Prompt framing effect:} An LLM generates different predictions depending on how prompts are framed. Predictions are associated with prompts instead of factual knowledge learned in LLMs. As shown in Figure~\ref{semantic-bias}, for a fact represented in a triplet \emph{$<$Cunter, is located in, Switzerland$>$}, the generated predictions for re-framed prompts \emph{``Which country is Cunter situated?''} and \emph{``Cunter is located in Switzerland. True or False?''} are non-factual.

\paragraph{Effect of in-context interference:} An LLM leverages in-context information during its decoding stage. The in-context information is concatenated with a test input (prompt) and acts as a condition when inferring hidden states concepts \citep{bay,bay1}. In-context information may negatively affect an LLM's prediction during knowledge probing. As shown in Figure~\ref{leakage-bias}, for the same fact, when presented with a context \emph{``England.''} concatenated with the prompting question \emph{``Which country is the location of Cunter?''}, an LLM generates a non-factual prediction \emph{``England''}.



\begin{figure}[htbp]
    \centering
    \includegraphics[scale=0.45]{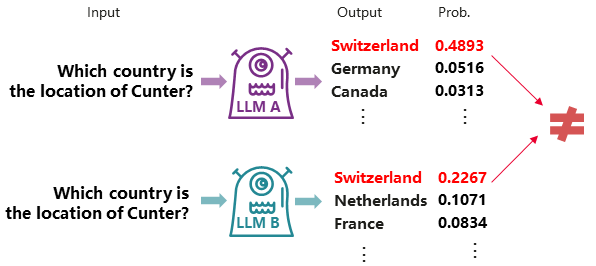}
    \caption{The same top-1 answer with different probabilities.}
    \label{question-mark}

\end{figure}
  
How do we assess the reliability of factual knowledge of LLMs under the effects of these hallucination-inducing factors? Investigations into the behaviors of language models during knowledge probing \citep{lama,negated,comprehension} have mainly used metrics like precision and accuracy to quantify errors under a specified factor like prompt framing \citep{cognitive} or mis-primed information \citep{negated}.  Despite the insights gained by showing the instability of LLMs during knowledge probing, these studies are subject to two limitations:

\begin{itemize}
    \item 

\textbf{No Exploration of Uncertainty}. Metrics like top-one accuracy may capture the ordering of predictions in the output space, but they lack the resolution to reflect on the degree of certainty of factual knowledge being learned by LLMs. Figure~\ref{question-mark} depicts an example where two LLMs (Models A and B) may produce the same result even though their output probabilities vary. By equating the performance of Model A with that of Model B, one introduces a level of approximation. This approximation in knowledge representation at the output space can be regarded as a source of uncertainty. In this paper, we directly use the output probabilities and construct a high-resolution metric to perform knowledge assessment. 
\item 

\textbf{Limited Scope}. Previous works focus on understanding the effect of variation of a specific type. We design experiments to investigate the combined effects of multiple causes of variation: prompt framing and in-context interference during knowledge assessment. In addition, few studies have experimented on LLMs with billions of parameters. In contrast, we investigate the knowledge reliability of 12 freely downloadable LLMs with a range of parameter sizes and origins (with and without instruction fine-tuning).\footnote{Only freely downloadable LLMs can be used as we need to use the output probability distributions.} 

\end{itemize}

In this paper, we propose a novel distance-based approach 
\textbf{MOdel kNowledge relIabiliTy scORe (MONITOR)} which captures the deviation of output probability distributions under contexts of prompting variance, and interference from mispriming \citep{negated} and positively-primed prompts. By leveraging the probability distribution of the output space, \textbf{MONITOR} serves as a high-resolution metric for assessing the reliability of factual knowledge of LLMs. 


We perform experiments 
on a comprehensive set of knowledge probing tasks and investigate the correlation between accuracy and MONITOR.
Through experiments with a large variety of different facts, we show that the proposed MONITOR has a significant correlation (
0.846 Pearson coefficient) with the average accuracy recorded in LLMs. 
Further analyses show that MONITOR can address the ``accuracy instability'' issue when used along with an end-to-end point measurement (like accuracy). Computing MONITOR takes only one-third GPU hours of those consumed by a comprehensive accuracy reliability study, making MONITOR a low-cost metric for assessing factual knowledge reliability of LLMs. We deploy MONITOR  on various factual knowledge probing tasks including question and answer (QA), word predictions (WP) and fact checking (FC).

\textbf{Our contributions are:}

\begin{enumerate}
    
    \item We design a novel LLM assessment method under various major hallucination-inducing factors using probability distributions from the output space. MONITOR is a high-resolution and low-cost metric suitable for evaluating the factual knowledge of LLMs under prompt framing effects and in-context interference;
      
    
    \item We construct the \textbf{FKTC} (Factual Knowledge Test Corpus) test set by developing QA probing prompts (210,158 prompts in total) based on 16,166 triplets of 20 relations from the TREx dataset \citep{trex}. We release \textbf{FKTC} to the public to foster research works along this line.

\end{enumerate}

\section{Related Work}

\citet{lama} demonstrated that factual knowledge can be directly extracted from language models without needing an external knowledge source. However, extracting knowledge (aka knowledge probing) from language models is error-prone due to various biases. For example, \citet{consistency} showed that the consistency of knowledge extracted is generally low when the same fact is queried with different prompts. Many works in prompt engineering attempt to automatically construct prompts outperforming manual prompts \citep{autoprompt,lpaqa,ape,zero}. It is argued that the decent performance of a language model is ascribed mainly to the application of these biased prompts \citep{guess}, in which ``better'' prompts are found to over-fit the answer distribution of the test set instead of reflecting on LLMs' generalization ability to predict factual knowledge.

To ensure that language models are hallucination-free, we need to look at other factors originating from in-contextual information. For in-context bias, \citet{negated,comprehension} showed that language models fail on most negated probes and are easily misled by misprimes added to the probing context. On the other hand, \citet{calibrate,reliable,understand} found the presence of context biases in few-shot probing results. The works mentioned above focused on pinpointing issues affecting LLMs' factual prediction. Few studies were motivated to develop evaluation approaches insensitive to the hallucination-inducing causes. Recently, \citet{semantic} presented a framework for evaluating the consistency of LLMs based on accuracy. \citet{bench} designed a benchmark for assessing the robustness of LLMs to adversarial instruction attacks, measuring the corresponding end-to-end performance drops. \citet{aliases} proposed a new metric to measure factual knowledge capability under the bias caused by aliases (alternative names for entities or relations) by reducing the effect of entity and relation aliases in the factual probing. 
Without tackling other factors like prompt framing effects and in-context interference (and their interactions), the scope of the study is limited. The data is not released to the public, therefore a comparative analysis is not possible.

\section{LLMs under the Influence of Hallucination-Inducing Causes}
In this section, we investigate LLMs' accuracy under the influence of various hallucination-inducing causes mentioned above. We design five formats of prompts to demonstrate two categories of hallucination-inducing causes during knowledge probing (Table~\ref{probe}). Twelve LLMs with a wide range of parameter size (from 560 million to 30 billion parameters) are covered in this study and experiments (in Section~\ref{expriments}), including foundation language models of OPT \citep{OPT}, Galactica \citep{gala}, and instruction finetuned language model of BLOOMZ \citep{BLOOMZ}, Vicuna \citep{vicuna}, Flan-T5 \citep{t5},WizardLM \citep{wizard}, Flan-UL2 \citep{flanul2,ul2}, LLaMa-30b-instruct-2048 \citep{llama30}.

\begin{table}[] \scriptsize
\centering
\begin{tabular}{l}
\toprule
\textbf{Prompt frames}                                                            \\ \hline
(1) WP: {[}X{]} is located in \_                                                        \\
(2) QA: Which country is {[}X{]} situated?                                             \\
(3) FC: Statement: {[}X{]} is located in {[}Y{]}. The statement is True of False?      \\ \hline
\textbf{In-context interference}                                                   \\ \hline
(4) {[}Y{]}. Which country is the location of {[}X{]}? \\
(5) {[}Y\_{]}. Which country is the location of {[}X{]}? \\ 
\bottomrule
\end{tabular}
\caption{\label{probe} Examples of probing task templates extending P17 (the test dataset containing 931 subject-object pairs with the ``country'' relation from TREx \citep{trex}), where [Y] is the object corresponding to the subject [X], [Y\_] is an entity that has a weak relation with [X]. WP, QA, and FC stand for ``word prediction'', ``question and answer'' and ``fact checking''.}
\end{table}

\subsection{Effect of Prompt Framing on Accuracy}

Factual knowledge in masked language models (MLMs) is evaluated using cloze-style prompts to probe whether the model accurately predicts the masked token (i.e., ``object'' in ``subject-relation-object'' triplets). LLMs have no such constraint in the token generation. Therefore, we design three probing templates to show the effect of prompt framing on LLMs, depicted below, and for each task, we use seven paraphrased prompts to ensure diversity:

\textbf{Word Prediction (WP) Template}: Given the ``subject'' and the prompt template, LLMs perform word prediction to complete the sentence, e.g., the template (1) in Table~\ref{probe}. When LLMs generate a sentence rather than an ``object'' (as a one-word token), we manually evaluate the predicted results to ensure their validity.

\textbf{Question-Answer (QA) Template}: In the QA template, question prompts are constructed from manually paraphrasing templates in TREx \citep{trex} targeting each fact. For example, a template ``[X] is located in [Y].'' for a triplet <[X], is located in, [Y]> can be paraphrased to ``Which country is [X] situated in?''. 

\textbf{Fact Checking (FC) Template}: An FC prompt is designed as a verification statement based on a template in TREx, i.e., ``Statement: [X] is located in [Y]. The statement is True or False?''. We build the positive checking probe (\textbf{FC-pos}) and negative checking probe (\textbf{FC-neg}) corresponding to whether the statement is factual or not. For a negative fact-checking prompt, we average the prediction accuracy for the five random entities chosen from the same category weakly-related with the ``subject''.

The probing results are shown in Table~\ref{task} as accuracies in predicting P17 factual knowledge for each involved LLMs under prompting biases presented in terms of WP, QA, and FC templates. The performances of LLMs in predicting the fact test data vary significantly when presented with different prompting templates. 
The fluctuation under seven prompts shown as box plots in Figure~\ref{box} (Appendix~\ref{diversity}) further demonstrates the effect of prompt framing on the performance of LLMs.  

\begin{table}[htbp]\scriptsize
\centering
\begin{tabular}{l|l|l|llll}
\toprule
LLMs & Ins. & Size   & WP & QA   & \begin{tabular}[c]{@{}l@{}}FC-\\ pos\end{tabular} & \begin{tabular}[c]{@{}l@{}}FC-\\ neg\end{tabular} \\ \hline
BLOOMZ-560m &   $ \surd $    & 0.56 & 14.73    & 26.09 & 28.77 &  73.78   \\
BLOOMZ-1b1  &   $ \surd $    & 1.1  & 14.96    & 28.29 &  \textbf{0.11} &  \textbf{99.89}   \\
Galactica-1b3 &   $ \times $ & 1.3 & 2.36 & 46.43 & 86.05 & 12.29  \\
OPT-2b7     &   $ \times $   & 2.7  & 28.27    & 55.67 & 75.80 & 22.07  \\
BLOOMZ-3b   &   $ \surd $    & 3    & 20.46    & 30.69 & 58.29 & 81.95  \\
Vicuna-7b   &   $ \surd $    & 7  &  \textbf{34.89}   & \textbf{73.25 } & 91.19 & 85.67  \\
BLOOMZ-7b1  &   $ \surd $    & 7.1  & 26.26    & 33.72 & 88.32 & 64.98  \\
Flan-T5-XXL &   $ \surd $    & 11   & 51.47    & 31.01 & 88.05 & 78.78 \\
Vicuna-13b  &   $ \surd $    & 13   & 38.96    & 78.15 & 90.87 & 89.68 \\
WizardLM-13b &   $ \surd $    & 13 & 34.66 & 78.55 &87.71 & 93.89 \\
Flan-UL2    &   $ \surd $    & 20   & 21.57    & 46.44 & 79.51 & 73.58   \\
LLaMa-30b-ins. & $ \surd $   & 30   & 67.94    & 87.72 & 96.99 & 86.69  \\    
\bottomrule
\end{tabular}
\caption{\label{task} Accuracy of various LLMs in predicting factual knowledge of P17 relation. ``Ins.'' means whether the LLM has been instruction finetuned. The performances of LLMs have undergone significant variations for different prompting templates. The unit of ``size'' is billion. Abnormal performances of LLMs between QA and WP template-based probes (bold numbers of Vicuna-7b) and between the FC probes for positive and negative interference (bold numbers of BLOOMZ-1b1) are strong evidence of prompt framing effects. }
\end{table}

\subsection{Effect of In-context Interference}

To explore the effect of in-context interference bias,  we add probes with misprimed \cite{negated} interference by concatenating contexts in terms of factual/nonfactual information preceding the associated QA templates (shown in Table~\ref{probe}). Table~\ref{qa-interference} captures the accuracies of LLMs in a comparative study using factual entity probes and misprimes consisting of weakly associated entities. We observe strong interference effects from nonfactual antecedents for all 12 LLMs in our study. It can be observed that a factual entity (positive interference) can improve the accuracy by up to +43.67 while a weakly related entity information (negative interference) reduces the accuracy by -56.54 at most.

\begin{table}[]\tiny
\centering
\begin{tabular}{l|ccc}
\toprule
context &  $ \times $  & [Y]   &[Y\_]     \\ \hline
BLOOMZ-560m  & 25.91    & 66.17 (+40.26) & 14.50 (-11.41)   \\
BLOOMZ-1b1   & 27.74    & 64.02 (+36.28) & 16.99 (-10.75)  \\
Galactica-1b3      & 53.81    &  56.39 (+2.58) & 10.42 (-43.39)  \\
OPT-2b7      & 58.00   & 77.23 (+19.23)  & 19.83 (-38.17)\\
BLOOMZ-3b    & 35.38    & \textbf{79.05 (+43.67)} & 24.30 (-11.08)  \\
Vicuna-7b    & 73.25   & 99.67 (+26.42) & \textbf{16.71 (-56.54)}   \\
BLOOMZ-7b1   & 39.03    & 70.57 (+31.54) & 26.40 (-12.63)  \\
Flan-T5-XXL  & 37.85    & 42.53 (+4.68) & 29.77 (-8.08)   \\
Vicuna-13b   &  78.15  & 90.76 (+12.61)  &  44.58 (-33.57)    \\
WizardLM-13b   &  85.61  & 55.75 (-29.86) & 47.09 (-38.52)   \\ 
Flan-UL2     & 33.44    & 47.58 (+14.14) & 33.19 (-0.25)      \\
LLaMa-30b-ins. & 87.72 & 99.46 (+11.74) & 47.78 (-39.94)  \\
\bottomrule                                          
\end{tabular}
\caption{\label{qa-interference} The effect of probing for P17 relation with the QA template corresponding to the (4) and (5) in Table~\ref{probe}, where ``$ \times $'' means experimental results with the original QA templates, ``[Y]'' means results using the factual information as in-context information, and ``[Y\_]'' refers to results using non-factual in-context information of entities weakly related to ``[X]''.}
\end{table}

\section{MOdel kNowledge relIabiliTy scORe (MONITOR)}

\begin{figure}[htbp]
    \centering
    \includegraphics[scale=0.45]{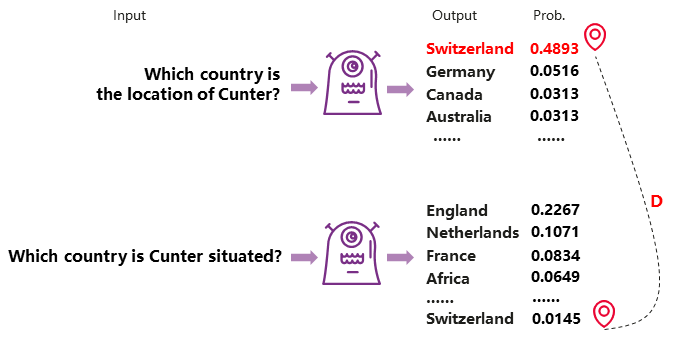}
    \caption{A primary anchor (for example, ``Switzerland'' with a probability of 0.4893) corresponds to its multiple foreign anchors with different output probabilities (i.e., ``Switzerland'' with a probability of 0.0145) when an LLM is exposed to different prompts. ``D'' refers to the distance measurement between the probabilities of two anchors.}
    \label{anchor}

\end{figure}

In this section, we introduce MONITOR, a distance-based score, to assess the factual knowledge of LLMs under the influence of previously mentioned prompt framing and in-context interference. 

An important notion of \textbf{``anchor''} is defined to establish a reference point, which is the valid factual knowledge represented as the answer probabilities in the output space. By calculating the distance (using the probability changes) between an anchor under investigation (known as the primary one) and its corresponding counterparts (aka the foreign anchors) in an influenced output space, we can measure how reliable an LLM is for the fact test set experimented. In order to enforce that the primary anchor is factually-accurate, we concatenate the correct answer preceding the associated QA template (e.g. Template 4 in Table~\ref{probe})\footnote{Adding positive in-context information to a prompt enhances the likelihood for an LLM to produce a correct answer. When none of the prompts works, the triplet is not included in the experiment.}, and the foreign anchors are generated using Templates 2 and 5 presented in Table~\ref{probe}. The distance calculation here fundamentally differs from that in ~\citet{aliases}, who leveraged a division between an specified relation and other irrelevant relations. 




Firstly, we introduce a new variable ($i$) to represent hallucination-inducing in-context information into the initial knowledge representation triple $<$\emph{subject, relation, object}$>$. The newly formed knowledge representation quadruple can be expressed as $<s, r, o, i>$. The information $i$ can be further categorized into two variables: positive information $i^{+}$ for factual object entities and negative information $i^{-}$ representing expressions serving as a bias for identifying $s$. For example,  ``France'' is considered as an $i^{-}$ when acting as a noisy condition to negatively affect an LLM in predicting a desirable outcome $<$\emph{Eibenstock, is located in, Germany}$>$. Corresponding to an object, $P(o|s,r,i)$ is the probability of the model generating the object $o$ with the conditions of subject $s$, prompt framing expression $r$, and the in-context information $i$.



To quantify the effect of $i$ on LLMs, we establish a reference point by treating the valid answer as the primary anchor mentioned above. As top-1 output probability can be used \citep{div} to detect false factual knowledge, we use the top-1 output probability to implement anchors. A \textbf{primary anchor} (for example, ``Switzerland'' with a probability of 0.4893 in Figure~\ref{anchor}) is defined as the valid output of an LLM for a base probe, which is the prompting template without any add-on context information. A primary anchor has multiple \textbf{foreign anchors} with various output probabilities (i.e., ``Switzerland'' with a probability of 0.0145 in Figure~\ref{anchor}) when an LLM is exposed to different prompts. In order to enforce that the primary anchor is consistently factually-accurate, we set the top-1 answer of the input with positive information $i^{+}$ as the primary anchor with probability $P(o|s,r,i^{+})$ and check their validity with Exact Match.

MONITOR consists of two distance-based measurement components: Prompt-framing Degree (PFD) and Interference-relevance Degree (IRD).
\subsection{Prompt-framing Degree}

The prompt-framing degree (PFD) is the mean distance between the output probability distributions of an enforced-accurate result (primary anchor) and the output probability distributions produced by the same LLM using prompting frames probing the same fact (foreign anchors). PFD evaluates the similarity of two output probabilities between prompting frame relation expressions $r$ (the basic prompt framing) and $r_j$. It is defined as:

\small 
\begin{gather} 
    PFD =  \frac{1}{R} \sum_{j=1}^R \frac{1}{ L_c }\sum_{l=1}^{L_c} {|P(o_c|s_c, r, i^{+})_{l} - P(o_c|s_c, r_j)_{l}|}
    \label{pfdeq}
\end{gather}
\normalsize where $R$ is the count of prompt framing expressions for a subject, and the count of
subject and object in a factual relation is $S$, $ c\in  \big\{1,...,S\big\}$. $L_c$ is the length of the anchor in terms of the number of subwords in the $c$-th object. PFD is a cumulative metric for assessing an LLM's capability in producing output probability distributions sharing the same characteristics under various prompting frames. PFD has a value between $0$ and $1$. The smaller the value is, the more robust an LLM is under the influence of prompt framing.

\subsection{Interference-relevance Degree}

Interference-relevance Degree (IRD) is the distance between the output probability distributions of accurate results enforced with positive information (primary anchor) and the probability distributions generated by the same LLM under the influence of in-context interference (foreign anchors). IRD measures an LLM's capability to predict factual knowledge under the influence of in-context interference.

\small \begin{gather} 
    IRD = \frac{1}{M} \sum_{m=1}^M  \frac{1}{ L_c }\sum_{l=1}^{L_c} {|P(o_c|s_c, r, i^{+})_{l} - P(o_c|s_c, r, i_{m}^{-})_{l} | }
\end{gather}
\normalsize 
We define the count of positive and negative information as one and $M$, respectively, corresponding to an object. IRD has a value between $0$ and $1$. As positive contextual information likely leads to factual knowledge generation, a smaller value of IRD indicates a lower level of influence from in-context interference biases.

\subsection{MONITOR: MOdel kNowledge relIabiliTy scORe}

The prompt-framing degree PFD and interference-relevance degree IRD are integrated to produce the proposed model knowledge reliability score  (MONITOR). MONITOR captures the quadratic interaction of PFD and IRD, as illustrated in Eq \ref{MONITOR} for a specified number of quadruples $<s, r, o, i>$, where the count of subject and object is $S$. A set of coefficients ($\alpha_{1-3}$) is introduced to quantify the contributions from PFD, IRD, and their interaction on MONITOR. In this experiment, we consider an equal contribution scenario ($\alpha_1 = \alpha_2 = \alpha_3 = 0.33$). The smaller the value of MONITOR, the less the model is influenced by hallucination-induced factors when producing factual outputs. Taking the average output probabilities of primary anchors for an LLM as the denominator, MONITOR captures the degree of knowledge learned by an LLM when assessing its factual knowledge. MONITOR measures the effects of prompt framing and interference per unit of average primary anchor probability, demonstrating the strength of anchor representations. 

LLMs are resource-hungry even during their inference phases. It is essential to ensure that an assessment metric is computation-efficient. Combining PFD, IRD, and their interaction in one metric can reduce the computation cost when evaluating factual reliability. Considering a relation with R prompt frames, M negative interference, and one positive interference, there are $R*M$ combinations required to compute the average accuracy (and accuracy range). In comparison, we only require $R + (1 + M)$ combinations to obtain MONITOR. The computation complexity for calculating MONITOR ($O(n)$) is considerably lower than that of accuracy ($O(n^2)$).

\tiny
\begin{gather}
    MONITOR = \frac{\sum_c^S{\sqrt{\alpha_1 PFD^2 + \alpha_2 IRD^2 + \alpha_3 PFD* IRD}}}{ \sum_c^S{\frac{1}{ L_c }\sum_{l=1}^{L_c}{ P(o_c|s_c, r, i^{+})_{l} }}}
    \label{MONITOR}
\end{gather} 
\normalsize
\section{Experiments}
\label{expriments}
In this section, we describe how to apply MONITOR to assess the factual knowledge of the 12 LLMs as mentioned above.

\subsection{Data Setting}
In this section, we describe how we develop a test corpus to accommodate prompts with various styles and in-context interference.


\textbf{Expanding Probing Prompt:} 
Based on 16,166 $<$subject, relation, object$>$ triplets from T-REx \citep{trex}, we develop QA probing prompts. We expand the probing prompt dataset by paraphrasing using GPT-4 \citep{gpt4} to create seven prompt frames for each triplet. In order to ensure the diversity of prompts, we choose prompts with a similarity score (BLEU) below a threshold (0.7).

\textbf{Adding In-context Interference:} Based on the QA prompts constructed above, we create a test dataset to explore the effectiveness of the designed metric with in-context interference biases. The dataset FKTC stands for ``Factual Knowledge Test Corpus''. 
Following the template patterns (Templates 4 and 5) in Table~\ref{probe}, we concatenate interference information (in terms of positive and negative in-context information) with the probing question for each subject. The negative information is entities from the same category weakly related to the corresponding subject, sampled from all objects that share the same relation. This process is applied to all templates presented in Table~\ref{example}, to produce 210,158 prompts focusing on 20 relations.

\begin{table}[htbp] \tiny
\centering
\begin{tabular}{l|ccccc}
\toprule
LLMs           & MONITOR $ \downarrow $     & avg$ \uparrow $  & max $ \uparrow $  & min $ \uparrow $ & probs $ \uparrow $ \\ \hline
BLOOMZ-560m    & 0.701 & 27.770  & 40.411  & 15.062  & 0.467        \\
BLOOMZ-1b1     & 0.692 & 30.055  & 43.369  & 16.654  & 0.501        \\
Galactica-1b3  & 0.747 & 22.936  & 39.414  & 9.427   & 0.637        \\
OPT-2b7        & 0.637 & 25.599  & 37.117  & 11.347  & 0.360        \\
BLOOMZ-3b      & 0.686 & 30.638  & 44.760  & 16.760  & 0.610        \\
Vicuna-7b      & 0.504 & 38.194  & 59.727  & 18.361  & 0.884        \\
BLOOMZ-7b1     & 0.632 & 36.232  & 49.328  & 22.870  & 0.613        \\
Flan-T5-XXL    & 0.630 & 32.968  & 48.864  & 19.868  & 0.798        \\
Vicuna-13b     & 0.484 & 44.882  & 65.499  & 26.967  & 0.862        \\
WizardLM-13b   & 0.560 & \textbf{51.477}  & 66.036  & \textbf{33.076}  & 0.774        \\
Flan-UL2       & 0.684 & 32.723  & 51.442  & 16.319  & 0.711        \\
LLaMa-30b-ins. & \textbf{0.479} & 50.798  & \textbf{71.188}  & 30.516  & \textbf{0.909}       \\ \hline
 Correlation& \multicolumn{2}{c}{Pearson}& & \multicolumn{2}{c}{p-value}\\ \hline
 r(MONITOR,avg acc)& \multicolumn{2}{c}{\textbf{-0.846}}& & \multicolumn{2}{c}{\textbf{0.001}}\\
 \bottomrule
\end{tabular}
\caption{\label{overall} The overall results are evaluated on FKTC with ``bold'' numbers indicating the best measurement over the same column category. The ``avg'', ``max'', and ``min'' mean the average, maximum, and minimum accuracy across the 20 test datasets. The ``probs.'' depicts the probabilities of primary anchors. ``$ \downarrow $'' means a smaller measurement wins. }
\end{table}

\begin{table*}[htbp]\scriptsize
\centering
\begin{tabular}{l|c|lll|lll|lll}
\toprule
\multirow{2}{*}{LLMs} & \multirow{2}{*}{Ins.} & \multicolumn{3}{c|}{P178}                                           & \multicolumn{3}{c|}{P108}                                           & \multicolumn{3}{c}{P37}                                            \\
                      &                             & MONITOR $ \downarrow $ & avg acc  $ \uparrow $ & probs.  $ \uparrow $ & MONITOR $ \downarrow $ & avg acc  $ \uparrow $ & probs.  $ \uparrow $ & MONITOR $ \downarrow $ & avg acc  $ \uparrow $ & probs.  $ \uparrow $ \\ \hline
BLOOMZ-560m           & $ \surd $             & 0.594                           & 53.260                           & 0.471                & 0.947                           & 2.634                            & 0.313                           & 0.669                           & 33.142                           & 0.679                           \\
BLOOMZ-1b1            & $ \surd $             & 0.492                           & 56.752                           & 0.684                & 0.853                           & 7.454                            & 0.191                           & 0.662                           & 39.679                           & 0.751                           \\
Galactica-1b3         & $ \times  $           & 0.595                           & 27.763                           & 0.543                & 0.876                           & 0.686                            & 0.393                           & 0.639                           & 42.444                           & 0.703                           \\
OPT-2b7               & $ \times  $           & 0.470                           & 64.119                           & 0.348                & 0.739                           & 12.420                           & 0.343                           & 0.471                           & 52.866                           & 0.419                           \\
BLOOMZ-3b             & $ \surd $             & 0.624                           & 50.460                           & 0.863                & 0.858                           & 17.639                           & 0.436                           & \textbf{\textit{0.570}}                           & \textbf{\textit{51.242}}                           & 0.797                           \\
Vicuna-7b             & $ \surd $             & 0.339                           & 64.575                           & 0.969                & 0.620                           & 32.756                           & 0.969                           & \textbf{\textit{0.432}}                           & \textbf{\textit{51.384}}                           & 0.931                           \\
BLOOMZ-7b1            & $ \surd $             & 0.492                           & 60.865                           & 0.865                & 0.770                           & 31.340                           & 0.443                           & 0.462                           & 61.114                           & 0.827                           \\
FLAN-T5-XXL           & $ \surd $             & 0.368                           & 67.065                           & 0.852                & 0.676                           & 29.968                           & 0.855                           & 0.650                           & 34.773                           & 0.865                           \\
Vicuna-13b            & $ \surd $             & 0.327                           & 77.787                           & 0.955                & 0.632                           & 39.951                           & 0.899                           & \textbf{0.311} & 69.590                           & 0.942                           \\
WizardLM-13b          & $ \surd $             & 0.411                           & 84.878                           & 0.850                & 0.626                           & 54.735                           & 0.769                           & 0.467                           & \textbf{69.907} & 0.856                           \\
Flan-UL2              & $ \surd $             & 0.613                           & 49.968                           & 0.792                & 0.844                           & 23.942                           & 0.836                           & 0.575                           & 56.731                           & 0.738                           \\
LLaMa-30b-ins.        & $ \surd $             & \textbf{0.180} & \textbf{87.461} & \textbf{0.983}                & \textbf{0.522} & \textbf{60.493} & \textbf{0.972} & 0.411                           & 63.109                           & \textbf{0.950} \\ 
\bottomrule
\end{tabular}
\caption{\label{specific} Performance of various LLMs in predicting factual knowledge captured in P178, P108, and P37 \protect\footnotemark testing datasets with ``bold'' numbers indicating the winning measurement over the same column category. ``Ins.'' means whether the LLM has been instruction finetuned. The ``bold and italic'' fonts on P37 show how MONITOR can differentiate two LLMs (BLOOMZ-3b and Vicuna-7b) with a similar average accuracy.}
\end{table*}

\footnotetext{P178: the relation of developer. P108: the relation of employer. P37: the relation of official language.}

\subsection{Results}
\subsubsection{Overall Results}

The overall results are shown in Table~\ref{overall}, and the results of each relation are shown in Table~\ref{all-MONITOR} (Appendix), where MONITOR and the average accuracy (avg acc) are recorded for each LLM across the 20 test datasets in our experiments. Each LLM's minimal and maximal accuracy are also recorded to show the accuracy variability. MONITOR incorporates internal representations of an LLM (i.e., primary anchor probabilities) and the influences from various in-context biases on its representations (in terms of the proposed distance measurement among distributions).        

The proposed MONITOR can not only indicate the degree to which external inputs influence a model, but it also reflects on the strength of factual knowledge learned by taking account of average primary anchor probabilities across knowledge. As shown in Table~\ref{overall}, LLaMa-30b-ins stands out as the most capable (with the least MONITOR 0.479) LLM, followed by Vicuna-13b (0.484) and Vicuna-7b (0.504). MONITOR correlates significantly with the average accuracy (0.846 Pearson coefficient), indicating its suitability for evaluating factual knowledge of LLMs over large-scale test cases. 

As shown in Table~\ref{specific} (bold italic fonts), MONITOR can differentiate LLMs, for example, BLOOMZ-3b and Vicuna-7b, with a similar average accuracy on P37, by considering distance and probability information. We further discuss this in Subsection~\ref{MONITOR-acc}.  

It is worth noting that MONITOR adheres to the scale law via which a larger LLM tends to outperform smaller models in the same series (further in Subsection~\ref{llm-scale})   

\subsubsection{Results for Specific Facts}

We present a detailed view of the knowledge assessment of LLMs by drilling down into specific relation types. Unlike the overall results in the previous subsection, showing a general trend, the results disclosed here show more detailed insights. As shown in Table~\ref{specific}, the overall winning LLM (i.e., LLaMa-30b-ins.) can lose its edge in a particular relation type (P37). 

An LLM trained with instruction finetuning (i.e., BLOOMZ-3b) does not consistently outperform a foundation model with an equivalent amount of parameters (for example, OPT-2b7) on results presented in Table~\ref{specific}. 

\begin{table}[htbp]\scriptsize
\centering
\begin{tabular}{l|ccc}
\toprule
LLMs     & MONITOR $ \downarrow $ & base acc $ \uparrow $ & std $ \downarrow $  \\ \hline
Flan-T5-XXL                & 0.772                    & 51.713                       & 31.023                                 \\
OPT-2b7               & 0.536                    & 64.027                       & 12.087                                   \\
Flan-UL2               & 0.706                    & 67.029                       & 33.981                                 \\
\textbf{BLOOMZ-560m}   & \textbf{0.490}           & \textbf{70.888}              & \textbf{17.253}               \\
\textbf{BLOOMZ-1b1}    & \textbf{0.426}           & \textbf{71.932}              & \textbf{11.891}                \\
Galactica-1b3         & 0.659                    & 74.086                       & 26.576                                 \\
\textbf{BLOOMZ-7b}     & \textbf{0.472}           & \textbf{78.922}              & \textbf{19.252}                \\
\textbf{BLOOMZ-3b}     & \textbf{0.456}           & \textbf{79.143}              & \textbf{18.016}               \\
Vicuna-7b             & 0.427                    & 82.086                       & 27.585                                 \\
LLaMa-30b-ins.        & 0.543                    & 85.340                       & 34.131                             \\
\textbf{WizardLM-13b} & \textbf{0.425}           & \textbf{91.960}              & \textbf{8.978}                 \\
\textbf{Vicuna-13b}   & \textbf{0.190}           & \textbf{93.099}              & \textbf{5.768}                 \\ 
\bottomrule
\end{tabular}
\caption{When two LLMs with equivalent accuracies are assessed, an LLM with a lower MONITOR is likely to produce a lower standard deviation in accuracy. ``base acc'' is the accuracy associated with the base prompt. Bold fonts demonstrate evaluation cases.}
\label{std}
\end{table}

\begin{figure}[htbp]
    \centering
    \includegraphics[scale=0.5]{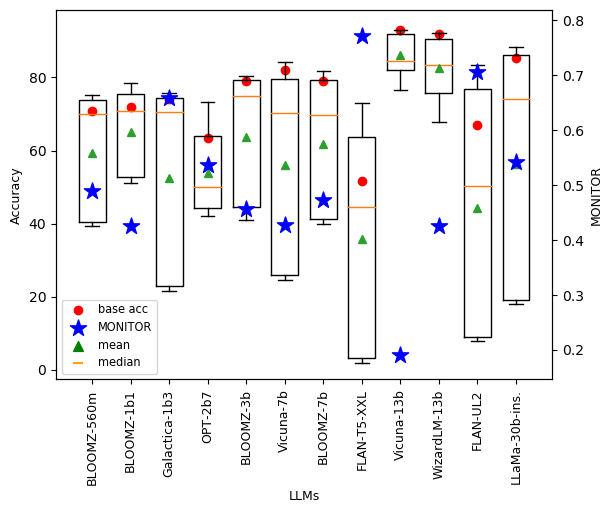}
    \caption{MONITOR can be used to differentiate LLMs' factual knowledge reliability when models with an equivalent base accuracy are evaluated. The box plots show the related distributions of accuracy. }
    \label{box-line}

\end{figure}

\section{Discussion}

\subsection{Accuracy Instability}

We analyze the LLMs' ``accuracy instability'' when predicting P1412 \footnote{P1412: the relation of languages spoken, written, and signed} with the results captured in Table~\ref{std} and Figure~\ref{box-line}. A variety of statistics, including the base accuracy (``base acc'') and standard deviation (``std'') of an LLM's accuracy, are recorded for comparisons. An LLM with a lower MONITOR has a lower accuracy standard deviation when two LLMs with equivalent base accuracies are evaluated. From an accuracy stability viewpoint, one may choose an LLM with a lower MONITOR. For example, we prefer Vicuna-13b over WizardLM-13b, even though they have similar accuracies as the MONITOR of Vicuna-13b is lower.

\subsection{MONITOR and Accuracy}
\label{MONITOR-acc}

\begin{table*}[htbp] \small
\centering
\begin{tabular}{l|lll|lll}
\toprule
              & \multicolumn{6}{c}{inputs}      \\ \hline
anchor        & \multicolumn{6}{l}{\textbf{French. }What language is the official language of Haiti?}      \\
in-context    & \multicolumn{6}{l}{\textbf{Irish. }What language is the official language of Haiti?}                                               \\
framing & \multicolumn{6}{l}{What language is considered the national language of Haiti?}   \\ \hline
              & \multicolumn{3}{c|}{output}           & \multicolumn{3}{c}{prob.}                  \\ \hline
              & anchor                                 & in-context                              & framing    & anchor    & in-context    & framing \\ \hline
BLOOMZ-3b     & French                                 & French                                  & French            & 0.761     & 0.411         & 0.527          \\
Vicuna-7b     & French                                 & French                                  & French            & \textbf{0.928}     & \textbf{0.622}         & \textbf{0.849}          \\ 
\bottomrule
\end{tabular}
\caption{Vicuna-7b outperforms BLOOMZ-3b in MONITOR when evaluated on P37 by producing correct answers with higher output probabilities in response to positive, negative in-context interference and prompt framing effect. }
\label{acc-probs}
\end{table*}

It can be observed in Table~\ref{overall} that the correlation between MONITOR and average accuracy is significant. How should one use MONITOR when assessing the reliability of LLM knowledge?     

\textbf{Resolution Characteristics:} We regard MONITOR as a high-resolution metric because it directly uses output probabilities and their changes (in terms of anchored distance) induced by hallucination factors. MONITOR considers both the output (nominal or qualitative data) and the probability of the output  (quantitative information). Comparatively, assessing LLMs' knowledge with an end-to-end metric, such as accuracy, is purely reliant on a nominal output from the softmax layer of a transformer. It is shown in Table~\ref{specific} that two LLMs (BLOOMZ-3b vs. Vicuna-7b) with almost identical average accuracy on P37 relation have two distinctive MONITORs (0.570 vs 0.432). Delving into the log file of the inference task, we gain in-depth insights into why Vicuna-7b outperforms BLOOMZ-3b in the reliability score. As shown in Table~\ref{acc-probs}, despite their similarities in the accuracy measurement, Vicuna-7b has much higher output probabilities than those of BLOOMZ-3b, contributing to their discrepancies in MONITOR.   

\begin{figure}[htbp]
    \centering
    \includegraphics[scale=0.35]{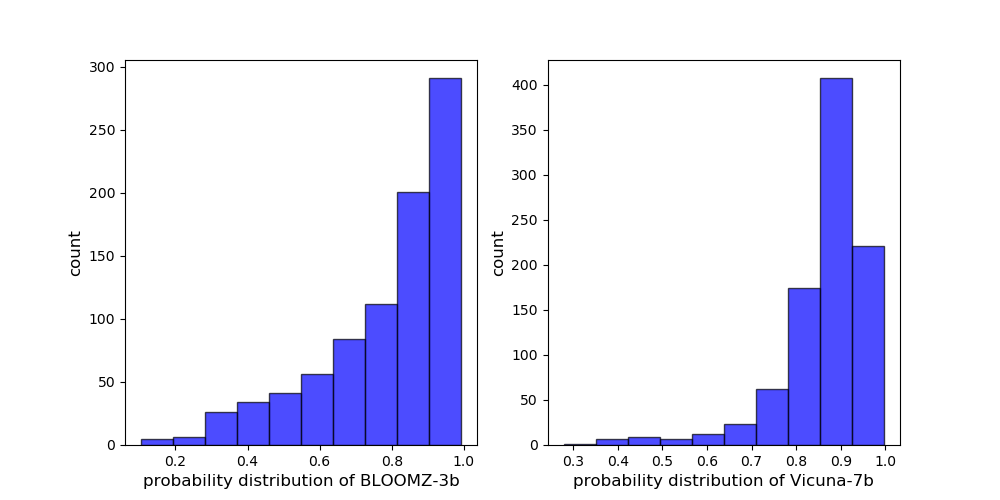}
    \caption{A comparison of the probability distribution of anchors between BLOOMZ-3b and Vicuna-7b on P37. The population percentages with a solid probability (aka, greater than 0.8) are \textbf{59\%} and \textbf{85\%} for BLOOMZ-3b and Vicuna-7b, respectively. } 
    \label{hist}
\end{figure}

Additionally, we plot out the probability distribution of the above two LLMs with almost identical average accuracy but very distinctive MONITOR (Figure~\ref{hist}). It can be observed that a more reliable LLM based on MONITOR, Vicuna-7b, has a much higher percentage of solid output probability than those of a volatile LLM (BLOOMZ-3b in this case). It is recommended to adopt MONITOR when using accuracy alone cannot differentiate LLMs' knowledge reliability.

\begin{figure}[h]
\centering                                                          
\includegraphics[scale=0.22]{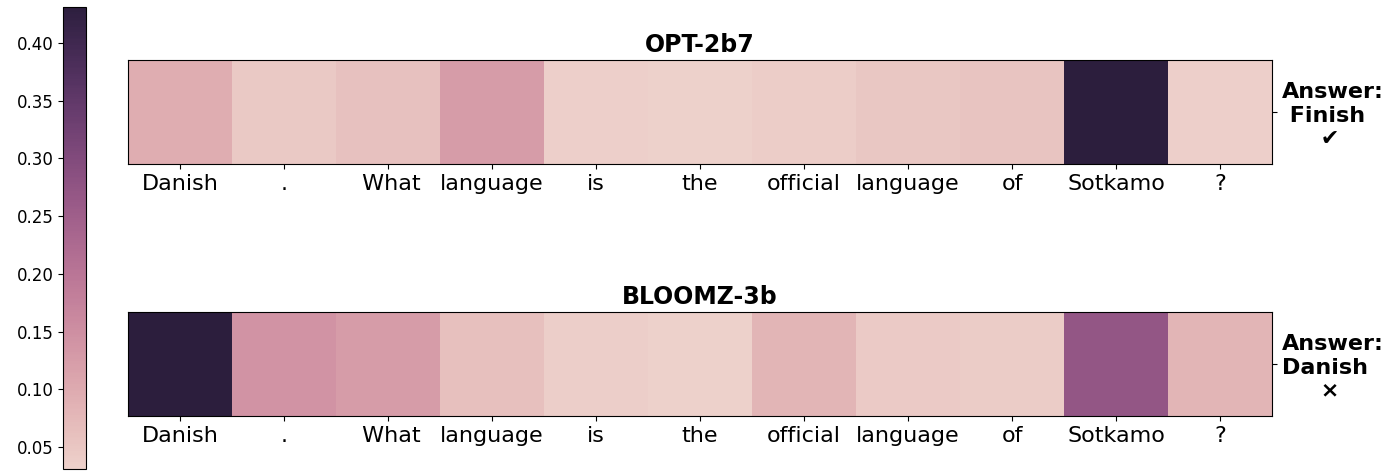}
\caption{Visualizing model behaviors of BLOOMZ-3b and OPT-2b7 under the influence of an input with misprimed in-context interference. The input is ``Danish. What language is the official language of Sotkamo?''. We evaluate the attribution of each input feature to the model's outputs by applying the integrated gradient technique.}
\label{vis} 
\end{figure}



 \textbf{Lower Computation Cost:} 

  \begin{table} \scriptsize
     \centering
     \begin{tabular}{l|cccc}
     \toprule
         Evaluation & MONITOR & Accuracy & MONITOR-saved \\ \hline
         GPU hours & 14.4 & 42.7 & 2.97X \\
    \bottomrule
     \end{tabular}
     \caption{GPU hours consumed calculating MONITOR and accuracy on a fact testset (P1412) of FKTC for LLaMa-30b-ins. ``MONITOR-saved'' denotes GPU hours saved from using MONITOR compared to accuracy. }
     \label{gpu}
 \end{table}
 
We compare the GPU hours consumed in producing MONITOR and a full-scale accuracy of LLaMa-30b-ins., which is experimented on a specific fact (P1412) test set using 8 NVIDIA V100 GPUs. It can be observed in Table~\ref{gpu} that using MONITOR leads to a 2.97-fold resource saving in GPU hours compared to applying an accuracy metric to a factual reliability evaluation. MONITOR is an economical method for assessing the reliability of LLM knowledge with scale. 

\subsection{Attribution of In-Context Interference} 
To demonstrate the resilience of LLMs with different MONITORs, we conduct an additional experiment by applying the Integrated Gradients~\citep{integrated} technique implemented in \citet{inseq}. By examining and visualizing the attribution of input features to the model's outputs, we can infer the reliability of LLMs with different MONITORs. We study the behaviors of two LLMs (OPT-2b7 vs. BLOOMZ-3b) with distinctive MONITORs (0.471 vs. 0.570). The heat map shown in Figure~\ref{vis} illustrates that a more reliable model with a smaller value of MONITOR, OPT-2b7, is less influenced by the in-context interference, producing a correct answer.


\begin{figure}[htbp]
    \centering
    \includegraphics[scale=0.45]{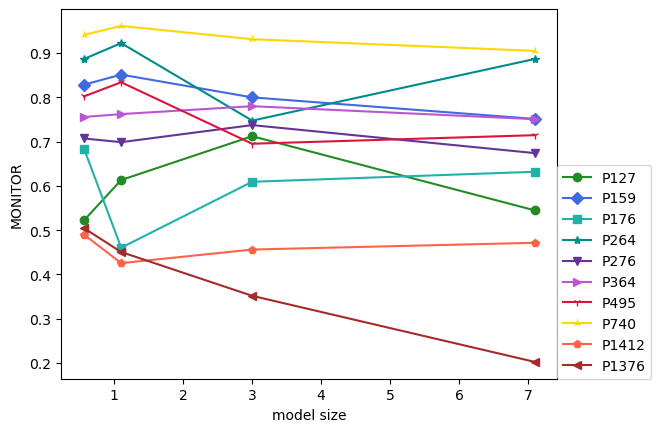}
    \caption{The BLOOMZ series adheres to the scale law for the specific facts with smaller MONITORs for bigger models. The horizontal axis represents the model’s size in billions, and the vertical axis represents the results of MONITOR.}
    \label{bloomz-scale}

\end{figure}

\begin{figure}[htbp]
    \centering
    \includegraphics[scale=0.5]{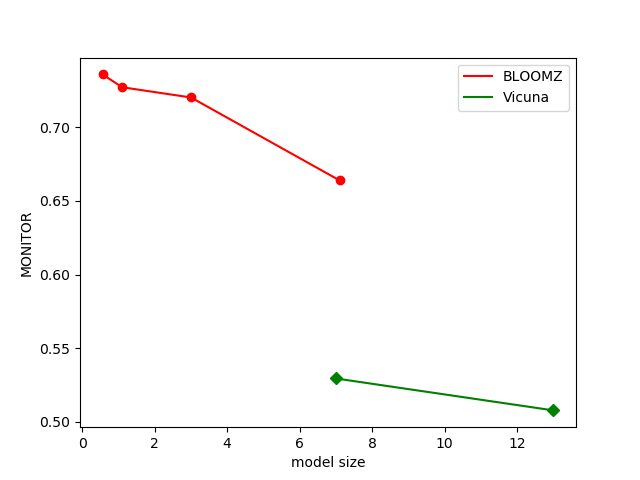}
    \caption{The BLOOMZ and Vicuna series adhere to the scale law based on the overall MONITOR results obtained from experiments on 20 test datasets. The horizontal axis represents the size of a model in billions, and the vertical axis represents the results of MONITOR.}
    \label{scale-all}

\end{figure}

\subsection{Analysis on LLMs Scale}
\label{llm-scale}
To further verify if MONITOR of LLMs follows the law of scaling, where larger LLMs are more knowledge-reliable, we present how MONITOR changes across BLOOMZ series for each specific fact (shown in Figure~\ref{bloomz-scale}). While MONITORs of LLMs may not conform to the scaling law at the granularity of each fact, their aggregated values in a comprehensive scope of experiments do follow the rule of scale (shown in Figures~\ref{bloomz-scale}-\ref{scale-all}). 

\subsection{Prompt Ablations}
\begin{figure}[htbp]
    \centering
    \includegraphics[scale=0.7]{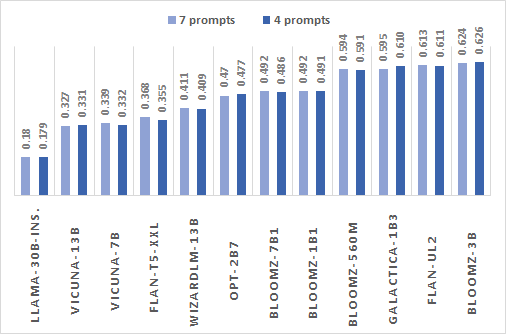}
    \caption{The consistency of MONITOR when assessing LLM's factual reliability in predicting P178 facts across different prompts settings.}
    \label{4prompts2}

\end{figure}

\begin{figure}[htbp]
    \centering
    \includegraphics[scale=0.7]{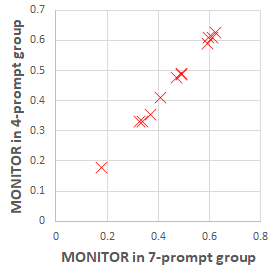}
    \caption{Significant correlation of MONITORs between the 7-prompt group and the 4-prompt group when assessing the reliability of LLMs in predicting facts from P178.}
    \label{4prompts3}

\end{figure}

We design an ablation study to investigate the consistency of MONITORs across different prompt settings by analyzing the MONITOR results in predicting P178 facts. The MONITORs from an expanded prompts group setting (consisting of seven prompts) and a sub-sampled group with four prompts are captured in Figures~\ref{4prompts2} and~\ref{4prompts3}. It is noted that MONITOR ranks LLMs in a consistent order for different prompt settings. Additionally, we observe a strong linear correlation between MONITORs of the expanded group and those from the sub-sampled group, indicating the scalability of MONITORs across prompt settings.    


\section{Limitation}

We focus on proposing MONITOR to assess the reliability of factual knowledge of LLMs during knowledge probing. Whether MONITOR can be generalized to a wider scope of tasks (e.g., summarization) warrants a future study. Additionally, the initial setup of contribution coefficients of PFD, IRD, and their interaction on MONITOR should be further investigated to establish an empirical benchmark. Currently MONITOR applies exact matching to obtain anchors to measure the reliability of LLM knowledge. Extending the automatic evaluation to anchors consisting of sentences is challenging.

\section{Conclusion}

In this paper, we show that large language models are subject to the influences of various hallucination-inducing causes. As a result, an end-to-end metric (like accuracy) is most likely to create an unstable reading. We propose a novel distance-based metric, directly measuring output probabilities and their changes to address ``accuracy instability'' caused by the prompt framing effect and in-context interference. A comprehensive scale of experiments demonstrates that the proposed MONITOR is a high-resolution economic method suitable for evaluating the reliability of large language model knowledge. The constructed FKTC dataset is available to the public to foster research along this line.  

\bibliography{anthology,custom}
\bibliographystyle{acl_natbib}

\newpage
\appendix

\section{Appendix}
\label{sec:appendix}

\begin{table*}[htbp]\tiny
\begin{tabular}{lllll}
\toprule
\#    & Relation                                         & Object Type              & Template                                     & prompt example                                       \\ \hline
P17   & country                                          & sovereign state          & {[}X{]} is located in {[}Y{]}.                       & Which country is the location of {[}X{]}?                  \\
P19   & place of birth                                   & city                     & {[}X{]} was born in {[}Y{]}.                 & Where was {[}X{]} born?                                    \\
P20   & place of death                                   & city                     & {[}X{]} died in {[}Y{]}.                     & In what place did {[}X{]} pass away?                       \\
P27   & country of citizenship                           & sovereign state          & {[}X{]} is {[}Y{]} citizen.                  & What country is {[}X{]} a citizen of?                      \\
P30   & continent                                        & continent                & {[}X{]} is located in {[}Y{]}.               & Which continent is {[}X{]} located in?                     \\
P37   & official language                                & language                 & The official language of {[}X{]} is {[}Y{]}. & What language is the official language of {[}X{]}?         \\
P101  & field of work                                    & organization             & {[}X{]} works in the field of {[}Y{]}.       & What is {[}X{]}'s area of expertise?                       \\
P103  & native language                                  & Indo-European languages  & The native language of {[}X{]} is {[}Y{]}.   & What is the native language of {[}X{]}?                    \\
P108  & employer                                         & business                 & {[}X{]} works for {[}Y{]}.                   & Which organization does {[}X{]} work for?                  \\
P127 & owned by & company & {[}X{]} is owned by{[}Y{]}. & Which company is the owner of {[}X{]}?\\
P159 & headquarters location & sovereign state & The headquarter of [X] is in [Y] . & In what city is [X] headquartered? \\
P176  & manufacturer                                     & manufacturer or producer & {[}X{]} is produced by {[}Y{]}.              & What is the manufacturer of {[}X{]}?                       \\
P178  & developer                                        & organisation             & {[}X{]} is developed by {[}Y{]}              & Which company is the creator of {[}X{]}?                   \\
P264 & record label & record label & {[}X{]} is represented by music label {[}Y{]}. &  What is the record label for {[}X{]}?\\
P276 & location & sovereign state  & {[}X{]} is located in {[}Y{]}. & What is the location of{[}X{]}?\\
P364  & original language of film or TV show             & Nostratic languages      & The original language of {[}X{]} is {[}Y{]}. & What is the native language of {[}X{]}?                    \\
P495  & country of origin                                & sovereign state          & {[}X{]} was created in {[}Y{]}.              & Which country was {[}X{]} created in?                      \\
P740 & location of formation & sovereign state  & {[}X{]} was founded in {[}Y{]}. & Which city was {[}X{]} founded in? \\
P1376 & capital of                                       & country                  & {[}X{]} is the capital of {[}Y{]}.                   & Which country's capital is {[}X{]}?                        \\
P1412 & languages spoken, written or signed              & Indo-European languages  & {[}X{]} used to communicate in {[}Y{]}.      & What language did {[}X{]} previously speak to communicate? \\
\bottomrule
\end{tabular}
\caption{\label{example} Examples of template for different relations and the corresponding prompts we build in this work.}
\end{table*}

\begin{table*}[htbp] \scriptsize
\centering
\begin{tabular}{l|cccccccccccc}
\toprule
knowledge     & \begin{tabular}[c]{@{}c@{}}BLOOMZ\\ -560m\end{tabular} & \begin{tabular}[c]{@{}c@{}}BLOOMZ\\ -1b1\end{tabular} & \begin{tabular}[c]{@{}c@{}}Galactica\\ -1b3\end{tabular} & \begin{tabular}[c]{@{}c@{}}OPT\\ -2b7\end{tabular} & \begin{tabular}[c]{@{}c@{}}BLOOMZ\\ -3b\end{tabular} & \begin{tabular}[c]{@{}c@{}}Vicuna\\ -7b\end{tabular} & \begin{tabular}[c]{@{}c@{}}BLOOMZ\\ -7b1\end{tabular} & \begin{tabular}[c]{@{}c@{}}Flan-T5\\ -XXL\end{tabular} & \begin{tabular}[c]{@{}c@{}}Vicuna\\ -13b\end{tabular} & \begin{tabular}[c]{@{}c@{}}WizardLM\\ -13b\end{tabular} & \begin{tabular}[c]{@{}c@{}}Flan\\ -UL2\end{tabular} & \begin{tabular}[c]{@{}c@{}}LLaMa-\\ 30b-ins.\end{tabular} \\
 \hline
P17   & 0.782     & 0.780     & 0.852    & 0.541   & 0.785    & 0.523    & 0.714     & 0.690      & 0.544     & 0.602       & 0.788   & 0.395         \\
P19   & 0.866      & 0.927     & 0.914        & 0.858  & 0.898    & 0.719    & 0.873     & 0.882      & 0.629     & 0.752       & 0.918   & 0.817         \\
P20   & 0.810      & 0.926     & 0.942        & 0.849  & 0.921    & 0.671    & 0.873     & 0.888      & 0.667     & 0.725       & 0.893   & 0.803         \\
P27   & 0.704      & 0.746     & 0.868        & 0.597  & 0.706    & 0.460    & 0.724     & 0.674      & 0.489     & 0.573       & 0.786   & 0.490         \\
P30   & 0.809      & 0.839     & 0.801        & 0.748  & 0.887    & 0.652    & 0.546     & 0.670      & 0.611     & 0.680       & 0.815   & 0.617         \\
P37   & 0.669      & 0.662     & 0.639        & 0.471  & 0.570    & 0.432    & 0.462     & 0.650      & 0.311     & 0.467       & 0.575   & 0.411         \\
P101  & 0.899      & 0.822     & 0.919        & 0.888  & 0.877    & 0.816    & 0.838     & 0.879      & 0.823     & 0.927       & 0.858   & 0.857         \\
P103  & 0.512      & 0.515     & 0.671        & 0.468  & 0.457    & 0.424    & 0.451     & 0.599      & 0.296     & 0.506       & 0.561   & 0.410         \\
P108  & 0.947      & 0.853     & 0.876        & 0.739  & 0.858    & 0.620    & 0.770     & 0.676      & 0.632     & 0.626       & 0.844   & 0.522         \\
P127  & 0.522      & 0.613     & 0.676        & 0.627  & 0.712    & 0.547    & 0.545     & 0.437      & 0.382     & 0.438       & 0.621   & 0.346         \\
P159  & 0.829      & 0.851     & 0.858        & 0.755  & 0.800    & 0.523    & 0.751     & 0.731      & 0.478     & 0.479       & 0.758   & 0.454         \\
P176  & 0.684      & 0.461     & 0.457        & 0.527  & 0.609    & 0.244    & 0.632     & 0.290      & 0.437     & 0.467       & 0.518   & 0.322         \\
P178  & 0.594      & 0.492     & 0.595        & 0.470  & 0.624    & 0.339    & 0.492     & 0.368      & 0.327     & 0.411       & 0.613   & 0.180         \\
P264  & 0.887      & 0.923     & 0.916        & 0.863  & 0.748    & 0.678    & 0.887     & 0.883      & 0.606     & 0.661       & 0.799   & 0.560         \\
P276  & 0.707      & 0.699     & 0.751        & 0.650  & 0.737    & 0.535    & 0.674     & 0.639      & 0.489     & 0.557       & 0.664   & 0.515         \\
P364  & 0.756      & 0.762     & 0.850        & 0.662  & 0.780    & 0.576    & 0.751     & 0.786      & 0.619     & 0.714       & 0.774   & 0.599         \\
P495  & 0.802      & 0.834     & 0.868        & 0.661  & 0.695    & 0.413    & 0.715     & 0.716      & 0.476     & 0.530       & 0.790   & 0.499         \\
P740  & 0.941      & 0.961     & 0.961        & 0.858  & 0.931    & 0.689    & 0.905     & 0.837      & 0.646     & 0.669       & 0.882   & 0.647         \\
P1376 & 0.505      & 0.451     & 0.606        & 0.602  & 0.352    & 0.299    & 0.202     & 0.158      & 0.501     & 0.555       & 0.202   & 0.079   \\
P1412 & 0.490      & 0.426     & 0.659        & 0.536  & 0.456    & 0.427    & 0.472     & 0.772      & 0.190     & 0.425       & 0.706   & 0.543         \\
\bottomrule
\end{tabular}
\caption{MONITOR for all involved LLMs experimented on FKTC dataset.}
\label{all-MONITOR}
\end{table*}

\begin{table*}
    \centering
    \begin{tabular}{l|cccccccccl}
    \toprule
         Pearson&  P17&  P19&  P20&  P27&  P30&  P37&  P101&  P103&   P108 &P127\\ \hline
         correlation &  -0.579&  -0.709&  -0.685&  -0.826&  -0.648&  -0.867&  -0.474&  -0.767&    -0.889&-0.926\\
         p-value &  0.048&  0.009&  0.013&  0.001&  0.023&  0.001&  0.119&  0.004&    0.001&0.001\\ \hline
 & P159& P176& P178& P264& P276& P364& P495& P740& P1376&P1412\\ \hline
 correlation & -0.941
& -0.941& -0.828& -0.950& -0.703& -0.740& -0.899& -0.919& -0.872&-0.900\\
p-value & 0.001& 0.001& 0.001& 0.001& 0.011& 0.006& 0.001& 0.001& 0.001&0.001\\
 \bottomrule
    \end{tabular}
    \caption{Pearson correlation between MONITOR and the average accuracy, evaluated on FKTC dataset.}
    \label{all-cor}
\end{table*}

\subsection{Prompt Framing Effect}
\label{diversity}
\begin{figure*}[h]
\centering                                                          
\subfigure[QA]{\label{17-qa}\includegraphics[scale=0.45]{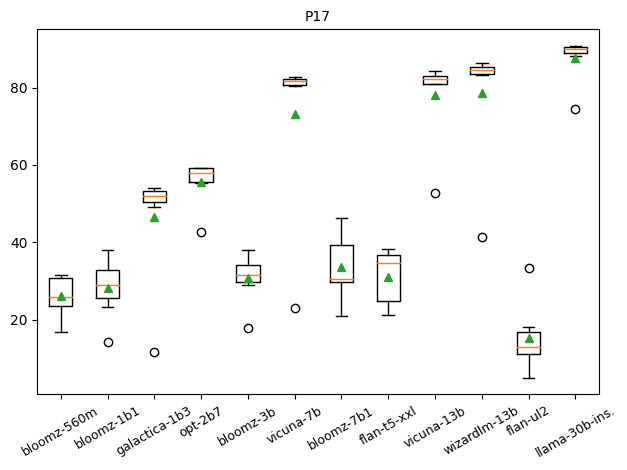}}        
\subfigure[WP]{\label{17-wp}\includegraphics[scale=0.45]{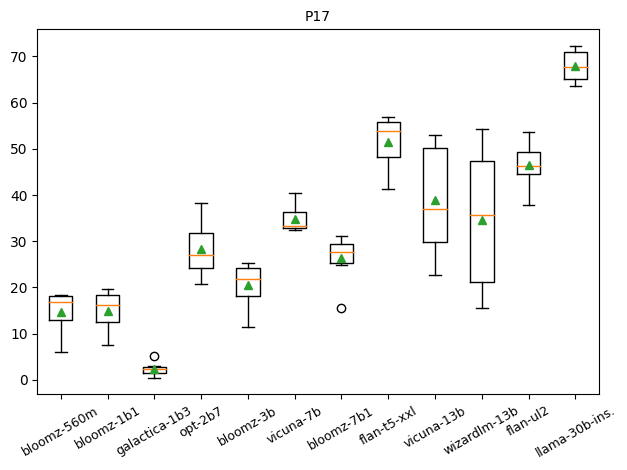}} 
\subfigure[FC-pos]{\label{17-fc-pos}\includegraphics[scale=0.45]{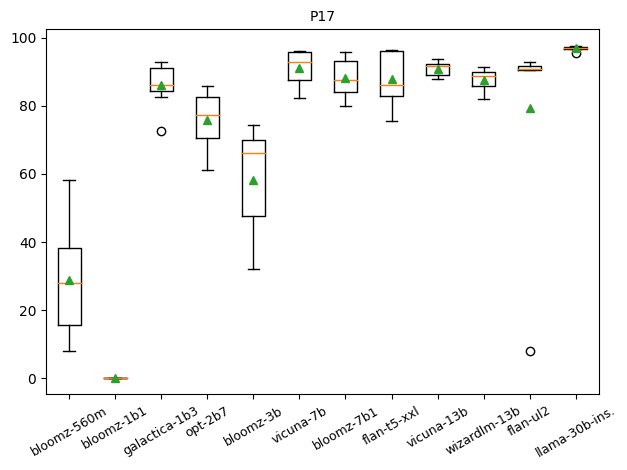}}
\subfigure[FC-neg]{\label{17-fc-neg}\includegraphics[scale=0.45]{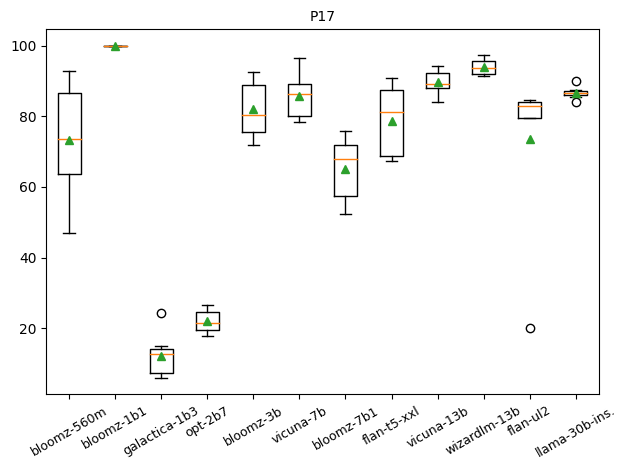}}        
\caption{\label{box} Box plots show the ``accuracy instability'' of LLMs under the effect of prompt framing in predicting P17 based on three tasks (WP, QA, and FC).}
\end{figure*}

Annotators paraphrase each relation in three prompting templates (WP, QA, and FC) so that each template can be used to produce seven prompts. For example, the template ``Which country is the location of [X]?'' could be paraphrased as: ``Which country is [X] situated?'', ``Which country can [X] be found?'', ``Which country is the geographical position of [X]?'', ``Which country is the site of [X]?'', ``In Which country is [X] situated?'', ``Whereabouts is [X] located?''. In this way, context diversity and semantic invariance are guaranteed. Figure~\ref{box} shows the ``accuracy instability'' of LLMs under the effect of prompt framing in predicting P17 facts based on three tasks (WP, QA, and FC).

\end{document}